\documentclass[journal]{IEEEtran}
\usepackage{amsmath,amsfonts}
\usepackage{ulem}
\usepackage{algorithmic}
\usepackage{algorithm}
\usepackage{multirow}
\usepackage{array}
\usepackage[table,xcdraw]{xcolor}
\usepackage[caption=false,font=small,labelfont=sf,textfont=sf]{subfig}
\usepackage{textcomp}
\usepackage{stfloats}
\usepackage{url}
\usepackage{verbatim}
\usepackage{graphicx}
\usepackage{cite}
\usepackage{hyperref}
\usepackage{amssymb}
\usepackage{booktabs}
\usepackage{color}

\begin{document}

\title{Lost in UNet: Improving Infrared Small Target Detection by Underappreciated Local Features}

\author{Wuzhou Quan, Wei Zhao, Weiming Wang, Haoran Xie,~\IEEEmembership{Senior Member,~IEEE}, Fu Lee Wang,~\IEEEmembership{Senior Member,~IEEE}, and Mingqiang Wei,~\IEEEmembership{Senior Member,~IEEE}
\thanks{W. Quan, W. Zhao and M. Wei are with the School of Computer Science and Technology, Nanjing University of Aeronautics and Astronautics, Nanjing, China, and also with the Shenzhen Research Institute, Nanjing University of Aeronautics and Astronautics, Shenzhen, China, (e-mail: q.wuzhou@gmail.com; weizhao0120@nuaa.edu.cn; mingqiang.wei@gmail.com).}
\thanks{W. Wang and F. L. Wang are with the School of Science and Technology, Hong Kong Metropolitan University, Hong Kong SAR (e-mail: wmwang@hkmu.edu.hk; pwang@hkmu.edu.hk).}
\thanks{H. Xie is with the School of Data Science, Lingnan University, Hong Kong SAR (e-mail: hrxie2@gmail.com).}
}

\maketitle

\begin{abstract}
Many targets are often very small in infrared images due to the long-distance imaging meachnism. 
UNet and its variants, as popular detection backbone networks, downsample the local features early and cause the irreversible loss of these local features, leading to both the missed and false detection of small targets in infrared images.
We propose HintU, a novel network to recover the local features lost by various UNet-based methods for effective infrared small target
detection.
HintU has two key contributions. First, it introduces the ``Hint" mechanism for the first time, i.e., leveraging the prior knowledge of target locations to highlight critical local features. Second, it improves the mainstream UNet-based architecture to  preserve target pixels even after downsampling. HintU can shift the focus of various networks (e.g., vanilla UNet, UNet++, UIUNet, MiM+, and HCFNet) from the irrelevant background pixels to a more restricted area from the beginning.
Experimental results on three datasets NUDT-SIRST, SIRSTv2 and IRSTD1K demonstrate that HintU enhances the performance of existing methods with only an additional 1.88 ms cost (on RTX Titan).
Additionally, the explicit constraints of HintU enhance the generalization ability of UNet-based methods.
Code is available at \url{https://github.com/Wuzhou-Quan/HintU}.
\end{abstract}


\section{Introduction}

\begin{figure*}[t]
    \centering
    \includegraphics[width=1\linewidth]{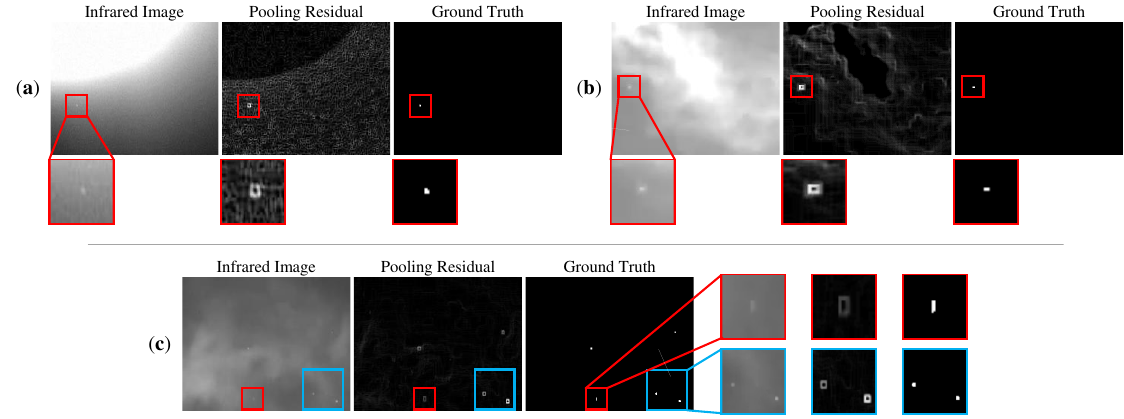}
    \caption{ \label{fig-glance}
        By visualizing the residual between the simple maximum pooling and the original image, we observe that the targets submerged in the high response gain a more distinct pattern.
        This newly acquired pattern is widely applicable, as evidenced by several visualizations portraying challenging scenarios.
        To augment the visual clarity, we have selectively magnified regions highlighted in \textcolor[RGB]{255,0,0}{red} and \textcolor[RGB]{0,176,240}{blue}.
        (\textbf{a})
        The target is submerged in low-contrast surroundings, with the image exhibiting a globally high response value difference.
        (\textbf{b})
        The distribution of image intensities appears chaotic, with an uneven distribution of strong response points.
        The target registers a response value lower than the global maximum.
        (\textbf{c})
        The global response is relative uniformity, featuring numerous targets with different degrees of distinguishability.
    }
\end{figure*}

\IEEEPARstart{I}{nfrared} imaging is a process where a thermal camera captures and creates an image of an object by using infrared radiation emitted from the object.
It has the ability to contactlessly and passively image from a distance in all-day or weather conditions.
Consequently, infrared small target detection (ISTD) has gained popularity and found wide applications in both civilian and military domains \cite{9604009}.

Infrared small target detection (ISTD) aims to find small targets from the infrared clutter background.
Despite numerous efforts, ISTD still faces many challenges like:
\begin{itemize}
    \item \textbf{Small Size}: Targets typically occupy only several to dozens of pixels in infrared images, due to the significant camera-to-object distance. In most ISTD datasets, more than 50\% of the targets have a size less than 0.05\% of the total number of pixels in input images~\cite{Jiang_2022_ACCV}.
    \item \textbf{Cluttered Background}: Substantial noise and background clutter always fill infrared images, making targets easy to obscure.
    \item \textbf{Unpredictable Type}: The target type is dynamic, with shapes and sizes varying across different scenes.
\end{itemize}

Traditional ISTD methods, such as filtering-based \cite{DetectionofDimTargets1996,maxmfilter}, low rank-based \cite{lowrank1,lowrank2}, and local contrast-based \cite{localcontrast1,localcontrast2}, have been proposed over the years.
However, the dynamic nature of real-world scenes complicates the effective utilization of handcrafted features and fixed hyperparameters.

Deep learning has dominated the field of ISTD~\cite{acm, uiunet, ifunetpp, ismallnet}.
For instance, Wang \textit{et al.}\cite{10.1117/12.2285689} are pioneers in migrating object classification techniques to ISTD.
However, most of them are directly transferred from the detection methods of natural images, and exhibit limited performance due to the tiny size of objects and the typical downsampling mechanism.

Recent efforts indicate that it is more effective to approach ISTD as a semantic segmentation problem rather than a typical detection problem.
This shift from detection to semantic segmentation may allow for more precise identification and localization of small targets, enhancing the overall detection performance.
For example, ACM \cite{acm} and ALCNet \cite{alcnet} individually build a bottom-up attention module and embedd the module into UNet~\cite{unet} and FPN in order to capture the targets’ saliency and discriminability characteristics.
Similarly, DNANet \cite{dnanet/nudtsirst} integrates more context features into UNet.
In another instance, UIUNet~\cite{uiunet} introduces more complicated intermediate states, effectively creating some beneficial underfitting.
This underfitting causes UNet to focus on shallow semantics, preventing overfitting and ensuring better generalization by concentrating on low-level features.
Additionally, ISNet~\cite{IRSTD1K} designs an independent branch to maintain the precise edge information.
ifUNet++~\cite{ifunetpp} lightweights the nested UNet structure and achieves better detection performance through cascading.

Both the network structures and downsampling strategies employed in the aforementioned UNet-based methods exacerbate the challenge of capturing minute details while risking the loss of local features.
For example, due to the dearth of high-level semantic information, small infrared target features are weakened in the deep layers of the CNN, which may underachieve the CNN's representation ability.
Efforts have been made to address these issues, including enhancing local feature extraction \cite{alcnet}, improving context understanding \cite{dnanet/nudtsirst}, and integrating global and local features \cite{mimistd}.
Nevertheless, these methods still fall short in adequately preserving and recognizing the local features for ISTD.

In this paper, we propose \textbf{HintU} to keep important low-level features of small targets from being missed by directly giving networks original shallow semantic information.
The potential of HintU lies in its ability to significantly enhance the performance of various ISTD methods that fail to adequately leverage this pattern.
HintU is easy to integrate into most UNet-like networks and widely adaptive to different datasets.
The contributions can be summarized as follows:
\begin{itemize}
  \item [1)]
  We take a fresh look at the ISTD task and rethink its natures, uncovering valuable clues that have been neglected for an extended period.
  
  \item [2)] 
  To address the neglection, we introduce the Hint mechanism.
  We implement such mechnism based on five existing UNet-like methods, including vanilla and advanced versions.
  Considering its adaptability to all UNet-like methods, we name the framework HintU.

  \item [3)]
    Extensive results confirm the effectiveness of HintU.
    It significantly boosts most of the ISTD methods, performing better than their vanilla versions.
    Besides, we discover some potential values of HintU and partially verify that HintU could bring a better generalization.
\end{itemize}

\section{Related Work}

\subsection{Infrared Small Target Detection}

In the early years, various traditional methods were proposed to detect small infrared targets.
For example, methods such as the top-hat filter \cite{tophat} and the max-median/mean filter \cite{maxmfilter} rely on manually designed filters to detect infrared small targets, but they suppress only uniform and simplistic background clutter.
Another category of methods, such as the local contrast measure \cite{localcontrast1}, the weighted strengthened local contrast measure \cite{localcontrast2}, \textit{etc.}~\cite{localcontrast3,localcontrast4,localcontrast5}, were inspired by the human visual system and focus on analyzing local contrast.
They excelled at differing high-contrast targets but struggled with dim ones.
Methods based on low-rank representation theory \cite{lowrank1,lowrank2} demonstrated adaptability to low signal-to-noise ratios but were easy to be distracted, particularly when dealing with extremely small or complex situations.
All the methods mentioned heavily depended on handcrafted features, proving inefficient and non-generalized.

The rapid evolution of deep neural networks has led to deep learning-based methods for ISTD that offer superior accuracy, generalization, and robustness.
Wang et al. \cite{10.1117/12.2285689} pioneered the migration of a model trained on the ImageNet Large Scale Visual Recognition Challenge (ILSVRC) to handle ISTD.
Similarly, Liu et al. \cite{liu2017image} proposed a solution based on a multi-layer perceptrion.
However, directly transferring methods from generic tasks often limits their performance in specialized domains like ISTD. 
Recently, researchers have developed specialized methods \cite{acm, alcnet, IRSTD1K, mdvsfa, 9199539,uiunet, ifunetpp,ismallnet,mimistd,hcfnet, rkformer} that significantly enhance performance.
For example, ACM \cite{acm} combined a new contextual module with a local contrast measure \cite{alcnet} to enhance performance. 
ISNet \cite{IRSTD1K}, inspired by the Taylor finite difference method, utilizes richer edge information to achieve refined, shape-aware detection.
RKFormer \cite{rkformer} introduces the Runge-Kutta ordinary differential equation method to Transformer architecture \cite{transformer}, exploring cross-level correlation to predict more accurately.
Then, MDvsFA \cite{mdvsfa} achieves a balance between false positive and false negative. 
UIUNet \cite{uiunet} introduces an interactive cross-attention nested UNet network, addressing the limitations of conventional attention mechanisms in ISTD.
MiM-ISTD \cite{mimistd} introduces the nested mamba architecture into ISTD. 
HCFNet \cite{hcfnet} offers an enhanced UNet tailored for small infrared object detection, addressing issues such as small target loss and low background distinctiveness.

\subsection{UNet}

UNet \cite{unet}, initially for medical image segmentation, utilizes an encoder-decoder architecture known for its low data volume requirements.
It comprises hierarchical downsampling steps followed by symmetric upsampling steps.
The skip connections link the downsampling and upsampling steps from level to level, ensuring semantic information preservation.
In recent years, various variants of it have been proposed.
For example, R2UNet \cite{r2unet} proposed recursive residual blocks in place of convolutional blocks.
Attention UNet \cite{attunet} utilizes attention gates within skip connections, selectively transmitting salient features while suppressing redundance.
UNet++ \cite{unet++} introduces nested middle layers to learn the projection of more complex features and semantics.
Although novel down-sampling and feature aggregating strategies have been proposed \cite{unet,unet++,attunet,r2unet,tridentnet}, the inherent mismatch with the ISTD task has still been ignored, which contributes to the decrease in performance.
Therefore, we focus on enhancing the generic UNet-like structure by incorporating useful prior information to fully unleash its potential.

\section{Method} \label{sec:method}

In this section, we present the motivation and implementation of HintU. 
In Subsection~\ref{subsec:motiv}, we introduce the motivation and formulation of HintU, and in Subsection~\ref{subsec:hint}, we introduce the implementation details of HintU.
In Subsection~\ref{subsec:unet}, we discuss the prototype networks, including the architecture resembling UNet, along with some noteworthy characteristics, followed by the loss function in in Subsection~\ref{subsec:loss}.

\begin{figure*}
    \centering
    \includegraphics[width=0.95\linewidth]{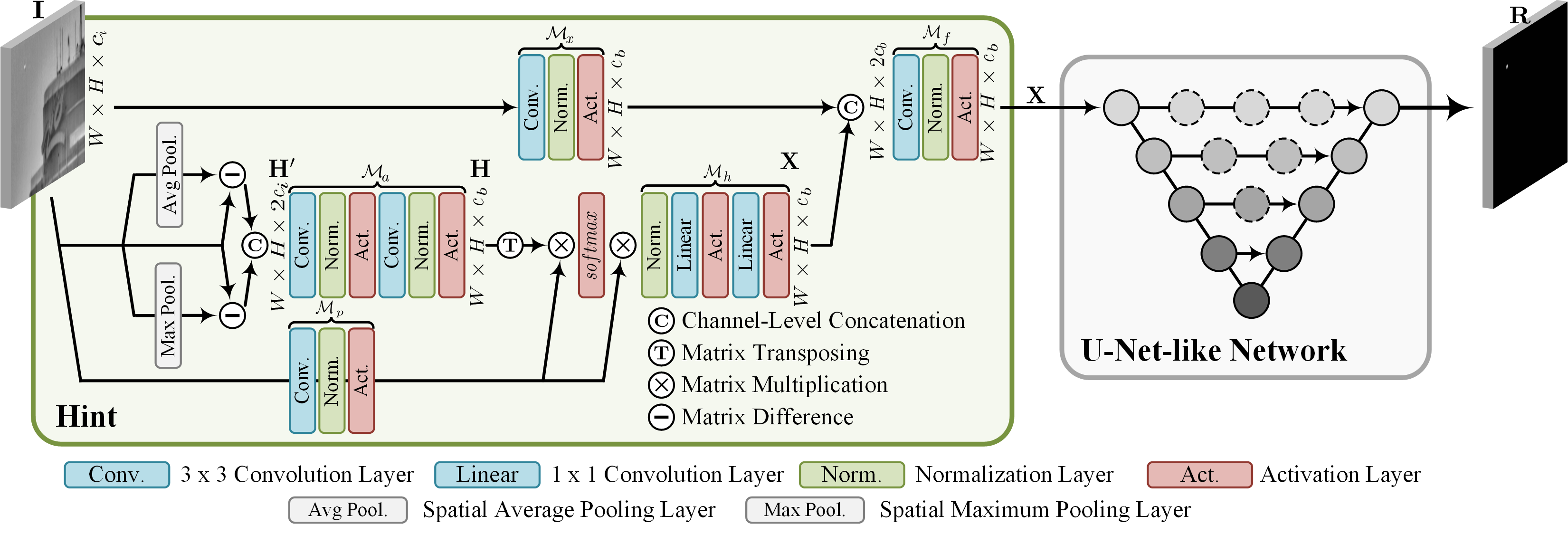}
    \caption{
    An overview of the proposed HintU for infrared small target detection.
    It consists of two primary modules: a prefixed Hint network, which generates representations based on hints derived from the original image, and a UNet-like network tasked with inferring the final results.}
    \label{fig-network}
\end{figure*}

\subsection{Motivation} \label{subsec:motiv}
Existing study \cite{lesps, deepimageprior} has revealed a phenomenon of the CNN in the training process. That is, it gradually converges the response values from the surroundings of the target to the ground truth.
In fact, it is part of how most attention mechanisms work: to provide the network with valuable responses beyond its field of view or training steps. Typically, this mechanism is employed in already-mapped feature spaces.
This is primarily due to the significance of objects in general image processing, which, while not always the main body of the image, are usually easily discernible.
Even following preprocessing and downsampling, their original semantic representations persist.

However, in infrared images, targets lack texture information and are notably diminutive, rendering any dimensionality reduction and abstraction detrimental to their original semantic patterns.
To achieve the effects similar to a complete attention mechanism in generic image processing, a significant increase in the number of parameters is required to adequately capture the imperceptible semantics in the new feature space \cite{dnanet/nudtsirst}.

Therefore, a new mechanism is required to operate on the layer as shallowly as possible, solely focusing on local semantics.
Consider the following scenario: all targets exhibit a response limited to only several pixels, often referred to as ``noise'' in common image processing.
The Laplace operator $\nabla^2$ can be formulated as:
\begin{equation}
    \label{eq-laplacine}
    \nabla^2u_s \propto \left(\frac{1}{(2r+1)^2}\sum \left\{u_p~|~p \in U(s,r)\right\}\right) - u_s ,
\end{equation}
where $u_s$ represents the response value at a specific location denoted by $s$, while $U(s,r)$ refers to the set of neighboring points within a radius $r$ from the central point $s$.
It emerges as an effective solution for mitigating this effect by quantifying the divergence of the spatial gradient field.
This calculation, in brief, represents the disparity between a point's value and the average of its neighboring pixels, swiftly suppressing the intensity of isolated points with a high response.
Typically, it ``shrinks'' the enclosed area.

In terms of the spatial size, the opposite operation $\mu$ can be formulated as:
\begin{equation}
    \label{eq-maxmu}
    \mu u_s \propto max(\left\{u_p~|~p \in U(s,r)\right\}) - u_s .
\end{equation}

Such opposite operation can extend the response area outside of the enclosed area (please refer to Figure~\ref{fig-glance} for visualization). From the visualizaton, it is observed that the tiny targets convoluted by the $\mu$ operator obtain a hollow neighborhood response with a wider range.

Then, for any region that is relatively small and has a higher response value than its adjacent pixels, the task shifts to finding a group of pixel clusters that gets dimmer or vanishes under the operator $\nabla^2u$ and has a fixed pattern under the operator $\mu$.
We magnify targets that are initially tiny (in terms of channels and space) to a scale that is more easily perceptible.
This hypothesis suggests that these two operators are more effective at identifying its presence.
Compared with the original image, the $\mu$ operator provides a larger perceptible range to avoid inadvertently losing key information.
It can be seen as a kind of nonlinear prior (for inference networks), expressed as:
\begin{equation}
    \label{eq-hprime}
    h^\prime_s := m(\nabla^2u_s,~\mu u_s) ,
\end{equation}
where $m(a,b)$ represents some kind of mixing strategy, such as addition, multiplication or vectorization.
For present, this is only a hypothesis that $h^\prime_s$ may make ultra-tiny targets more significant for a UNet-like network to perceive them.
The hypothesis will be implementated into a component in next subsection, validated in Subsection~\ref{subsec:hintonly}, and its effectiveness and principles analyzed in Subsection~\ref{subsec-principlesexp}.

\subsection{Hint} \label{subsec:hint}

Based on the hypothesis expressed in Equation~\ref{eq-hprime}, we build ``\textbf{Hint}'' to provide prior information to the network.
Hint serves as a plug-and-play component that operates before the original input $\mathbf{I}$ is processed by the backbone network $\mathcal{U}$.
The structure of Hint is illustrated in Figure~\ref{fig-network}.

First, for the original image $\mathbf{I} \in \mathbb{R}^{H \times W \times c_i}$, consisting of vectors $\mathbf{i}_{uv} \in \mathbb{R}^{c_i}$, where $u$ and $v$ denote the spatial location, we calculate the Hint priori tensor $\mathbf{H}^\prime$ point by point following Equation~\ref{eq-hprime}.
The channel-level concatenation is selected in HintU as the mixing strategy because the subsequent neural networks can be fitted to basic linear operations.
Then, the Hint prototype is projected by a group of convolutional blocks to determine the initial Hint tensor $\mathbf{H}$, expressed as:
\begin{equation}
\small
\label{eq-inihint}
  \left\{
  \begin{aligned}
    \mathbf{H} =& \mathcal{M}_a \circledast \mathbf{H}^\prime \\
    \mathbf{H}^\prime =& \left\{\mathbf{h}^\prime_{uv} | u \in [0, W), v \in [0, H)\right\} \\
    \mathbf{h}_{xy}^\prime =& \left(\mathbf{i}_{xy} - \frac{1}{k^2}\sum_{u=x-\lfloor\frac{k}{2}\rfloor}^{x+\lfloor\frac{k}{2}\rfloor}\sum_{v=y-\lfloor\frac{k}{2}\rfloor}^{y+\lfloor\frac{k}{2}\rfloor}\mathbf{i}_{uv}\right) \oplus \\
    & \bigg(\mathbf{i}_{xy} - max\Big(\left\{\mathbf{i}_{uv} | u \in \left[x-\lfloor\frac{k}{2}\rfloor, x+\lfloor\frac{k}{2}\rfloor\right], \right.\\
    & \left. v \in \left[y-\lfloor\frac{k}{2}\rfloor, y+\lfloor\frac{k}{2}\rfloor\right]\right\}\Big)\bigg)
  \end{aligned}
  \right.,
\end{equation}
where $k$ denotes the kernal size;
$\circledast$ denotes the application of a non-linear transformation;
$\oplus$ denotes channel-level concatenation;
$\mathcal{M}_a$ denotes a composite structure comprising two consecutive sets of convolutional blocks with distinct weights.
Each block encompasses a convolutional layer, a batch normalization layer, and a rectified linear unit (ReLU) activation layer.

Subsequently, a cross attention operation is conducted between the initial Hint tensor and original image to obtain the embedded Hint $\mathbf{H}_e$.
This operation efficiently maps the information contained in the Hint to the original data.
$\mathbf{H}_e$ is expressed as:
\begin{equation}
  \mathbf{H}_e = \mathcal{M}_h \circledast \left(softmax\left(\frac{\left(\mathcal{M}_p \circledast \mathbf{I}\right) \mathbf{H}^\mathrm{T}}{\sqrt{c_i}}\right)\left(\mathcal{M}_p \circledast \mathbf{I}\right)\right).
\end{equation}

Finally, $\mathbf{H}_e$ is fused with the original image I at the shallow semantic level to obtain the enhanced input $\mathbf{X} \in \mathbb{R}^{H \times W \times c_{base}}$, where $c_{base}$ denotes the latent space dimensions where $\mathbf{H}_e$ located, expressed as:
\begin{equation}
    \label{eq-x}
  \mathbf{X} = \mathcal{M}_f \circledast \left(\left(\mathcal{M}_x \circledast \mathbf{I}\right) \oplus \mathbf{H}_e\right).
\end{equation}

In order to facilitate evaluation, the $c_{base}$ used in all experiments is set to 32.
The enhanced input $\mathbf{X}$ then replaces the original I as input and is sent to the UNet-like network $\mathcal{U}$.

\subsection{UNet}  \label{subsec:unet}

\begin{figure*}
    \centering
    \includegraphics[width=0.95\linewidth]{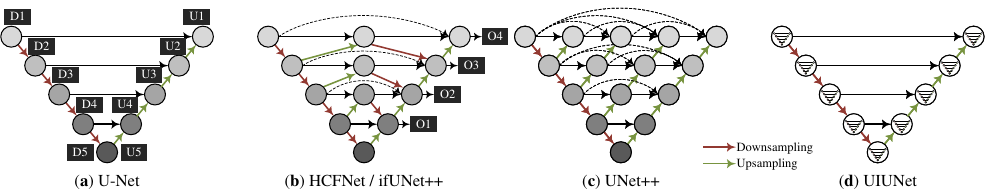}
    \caption{
    Several UNet-like network structures.
}
    \label{fig-unet-comp}
\end{figure*}

%While the structure of UNet-like networks is not the focal point, it is essential for elucidating the overall framework of HintU.
This section briefly describes several fundamental UNet-based methods utilized in our work.
Embedded Hint $\mathbf{X}$ will replace the usual original image $\mathbf{I}$ as the input of the UNet network.
Therefore, the UNet-like network in HintU needs to be slightly modified, that is, the number of input channels needs to be increased accordingly.
In the UNet family, semantic information is typically extracted from shallow to deep layers during the down-sampling process, while the up-sampling process expresses the semantics of interest.
However, traditional UNet networks undergo significant down-sampling, resulting in a drastic reduction in spatial resolution at each layer, halving it with each step.
Consequently, for many small targets, it is crucial that they are captured before the down-sampling reaches deeper layers.
Otherwise, it becomes challenging for them to reappear in the very deep layers and beyond.
This issue undermines the reliability of the prevalent jump links in UNet.
These links assume that information from the first and second layers has successfully captured the target.
Otherwise, even under optimal conditions, the jump link may convey information that lacks practical significance.

To demonstrate the frexbility of HintU, we employ vanilla UNet \cite{unet}, UNet++ \cite{unet++}, HCFNet \cite{hcfnet}, UIUNet \cite{uiunet}, and MiM-ISTD \cite{mimistd}.
The architectures of the four methods are shown in Figure~\ref{fig-unet-comp}.
For the MiM-ISTD which is designed for high-resolution ($512 \times 512$) image, we add some extra layers to fit downsampled images, named MiM+.

\subsection{Loss}   \label{subsec:loss}

Binary cross-entropy is hired to train our HintU as:
\begin{equation}
    \label{eq-bce}
    \delta_{bce}(y, \hat y) = -ylog\hat y - (1-y)log(1-\hat y),
\end{equation}
where $y$ and $\hat y$ denote the estimated results and the ground truth, respectively.
For HCFNet \cite{hcfnet}, we follow its deep supervision strategy as claimed.

\section{Experiments}

In this section, we discuss the experiment's details and results.
Firstly, in Subsection~\ref{subsec:exp_details}, we present the dataset, equipment platform, training parameter setup, and evaluation metrics employed in our experiments, \textit{e.t.c.}
In Subsection~\ref{subsec:hintonly}, we perform a small-scale experiment crafted to preliminarily validate that the Hint works under our expectations of equivalence rather than relying on any tricks.
In Subsection~\ref{subsec:comp_exp}, we execute extensive experiments across various method configurations on all datasets to affirm the effectiveness and universality of Hint.
Subsection~\ref{subsec-effexp} focuses on the costs of integrating the Hint.
Following that, in Subsection~\ref{subsec-principlesexp}, we discuss the deeper intrinsic principles of Hint, offering further analysis along with more visualization experiments.
Lastly, within Subsection~\ref{subsec-generalizationexp}, we adopt another perspective to explore the potential value of Hint for cross-domain inference.

\begin{figure*}[t]
    \centering
    \includegraphics[width=0.9\linewidth]{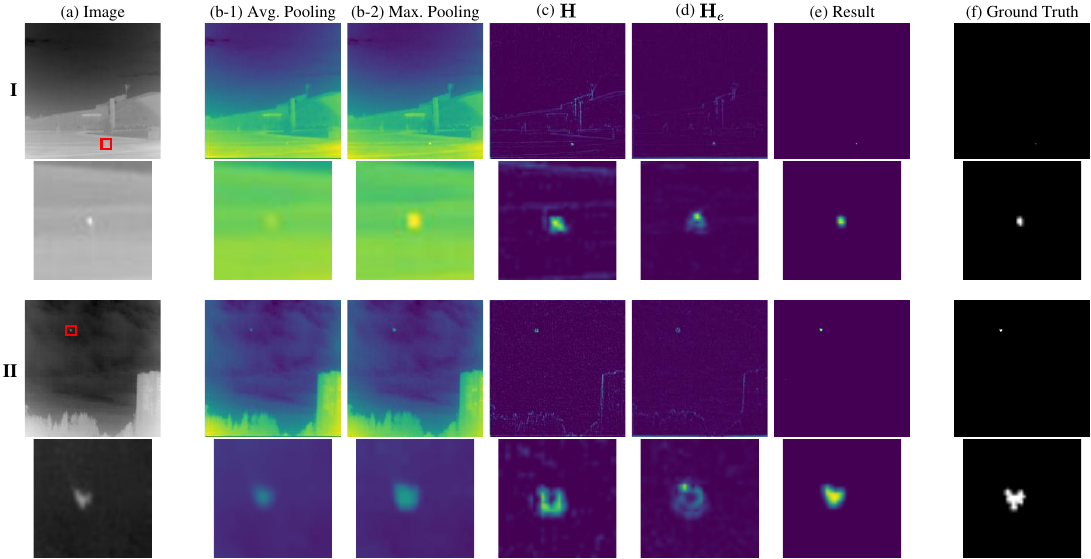}
    \caption{
    An overview of the HintO for infrared small target detection.
    In comparison to HintU, HintO completely excludes the original image when generating Hint representations, facilitating a direct assessment of the effectiveness of the Hint concept itself.
    For the enhanced visibility, highlighted areas are enlarged.
    }
    \label{fig:ab-vis-hinto}
\end{figure*}

\subsection{Datasets, Implementation Details, and Evaluation Metrics} \label{subsec:exp_details}
\begin{figure*}[t]
    \centering
    \subfloat[]
    {
        \begin{minipage}{0.3 \linewidth}
            {
                \label{fig-firstglance-1}
                \begin{center}
                    \centerline{\includegraphics[width=1\linewidth]{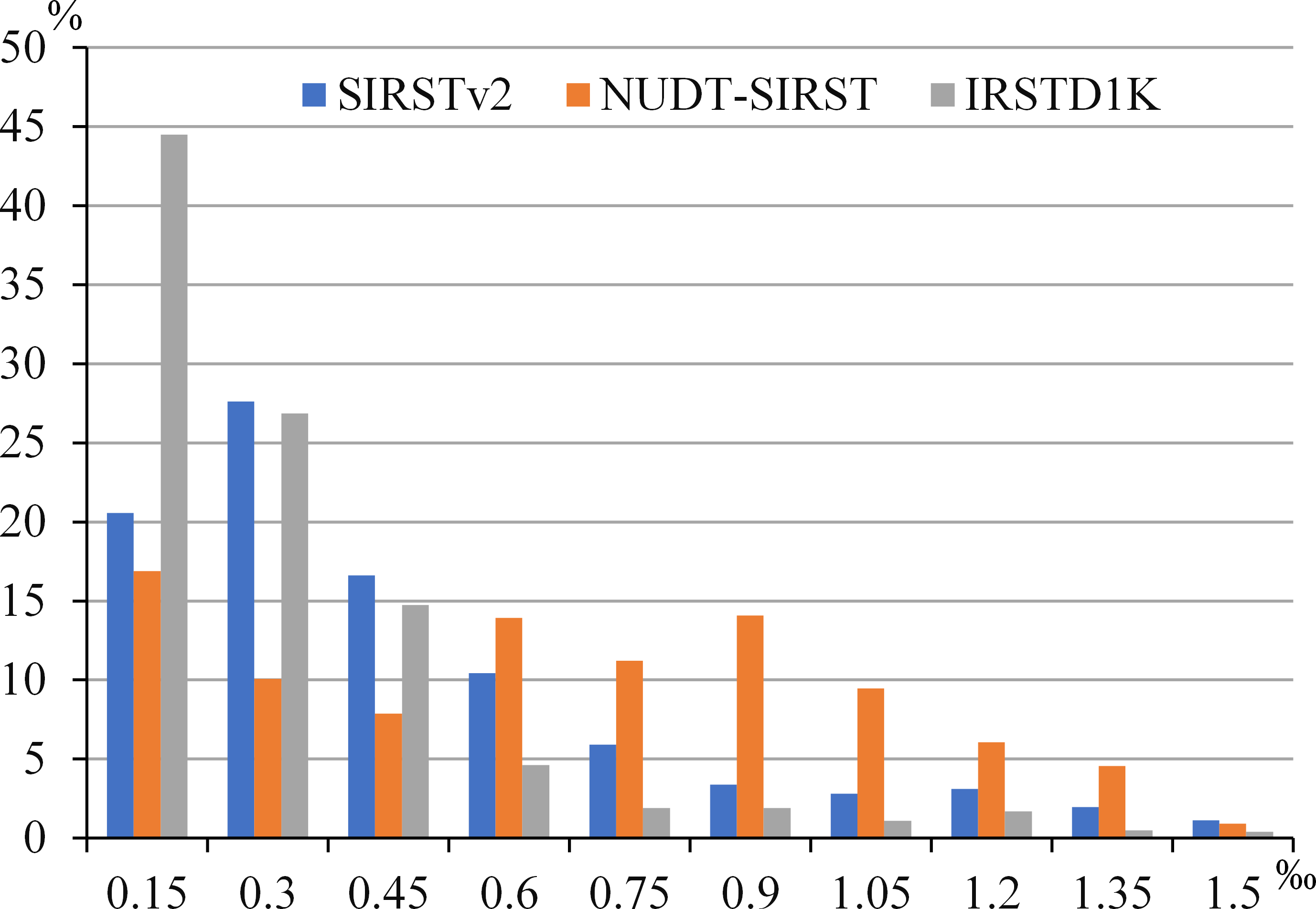}}
                \end{center}
            }
        \end{minipage}
    }
    \hspace{0.02 \linewidth}
    \subfloat[]
    {
        \begin{minipage}{0.3 \linewidth}
            {
                \label{fig-firstglance-2}
                \begin{center}
                    \centerline{\includegraphics[width=1\linewidth]{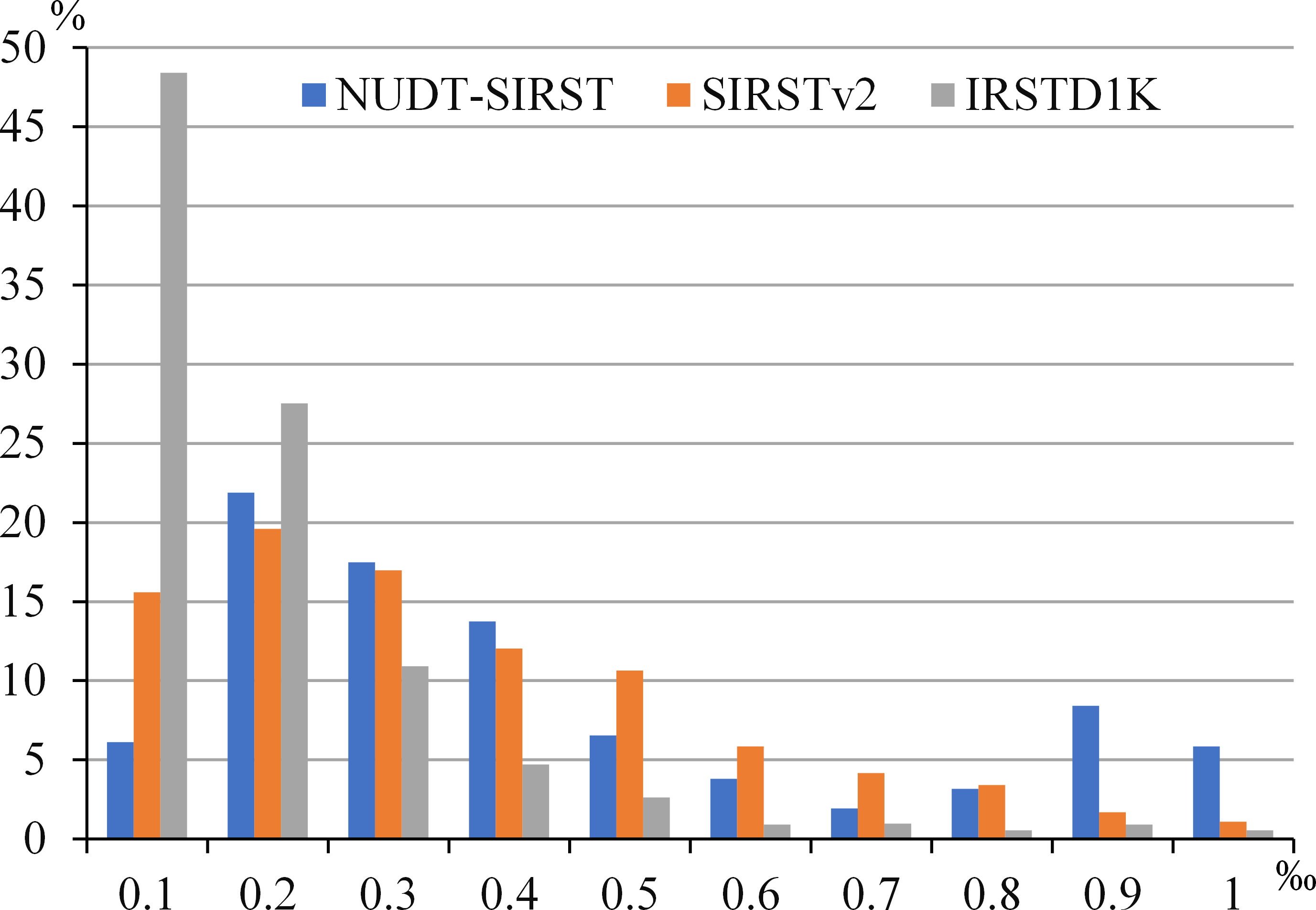}}
                \end{center}
            }
        \end{minipage}
    }
    \hspace{0.02 \linewidth}
    \subfloat[]
    {
        \begin{minipage}{0.3 \linewidth}
            {
                \label{fig-firstglance-2}
                \begin{center}
                    \centerline{\includegraphics[width=1\linewidth]{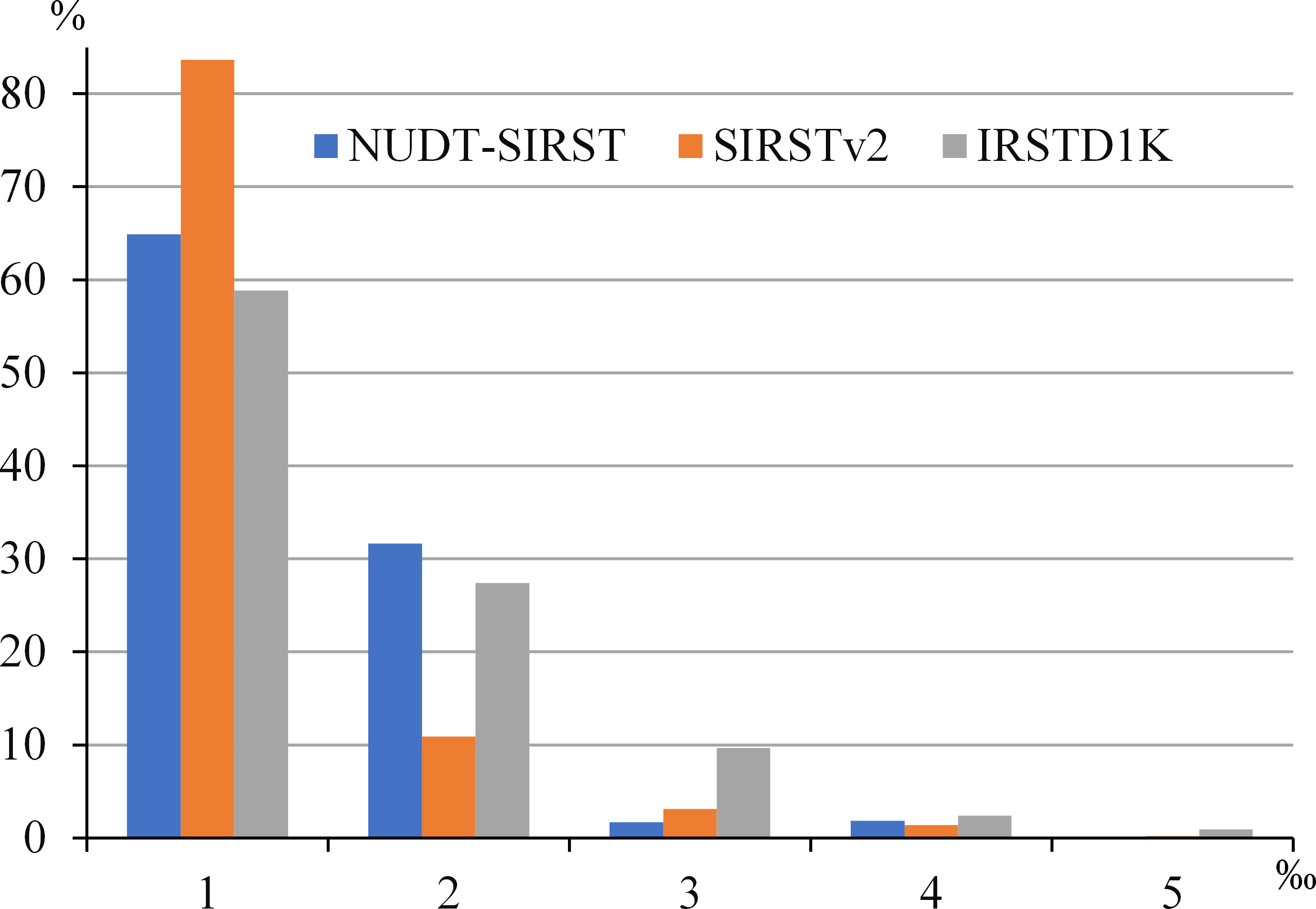}}
                \end{center}
            }
        \end{minipage}
    }
    \caption{ \label{fig:ds-size-dist}
        Summarized information of the three datasets,
        including (\textbf{a}) the ratio of the number of global high-response pixels to the total number of pixels
        (\textbf{b}) the ratio of the pixel count of each target to the total number of pixels
        (\textbf{c}) the count of targets in each frame.
        The x-axis denotes the value.
        The y-axis shows the percentage of total.
    }
\end{figure*}

We conduct experiments on three datasets, i.e., NUDT-SIRST \cite{dnanet/nudtsirst}, SIRSTv2 \cite{SIRSTv2}, and IRSTD1K \cite{IRSTD1K}.
Although they are all ISTD datasets, their difficulty in small target detection is different.
Among them, IRSTD1K is the least detectable data set for high-response pixel clusters. About 75\% of the targets only have less than 0.02\% of the frame size, and the average high-response points account for less than 0.03\%.
More specifically, taking $256\times 256$ resolution as an example, most of the target pixels are less than 13 ($3.6 \times 3.6$).
More detailed analysis are shown in Figure~\ref{fig:ds-size-dist}.
Given HintU is based on data-driven methods, we employ a 1:1 dataset division for training and testing, ensuring no overlap between them.

All the experiments are expressed on a machine with a single AMD® EPYC™ 7282 CPU and a single NVIDIA® TITAN RTX™ GPU.
We employ the AdamW \cite{adamw} optimizer with a learning rate of 1e-3 for network optimization.
The learning rate undergoes a gradual reduction using the cosine annealing algorithm, iteratively transitioning from 1e-3 to 1e-6.
The batch size is set to 8, and the target epoch number is set to 300.
The spatial resolution of all model inputs is standardized to $256 \times 256$, and likewise, the corresponding ground truth results are adjusted using the same strategy for loss calculation.
During evaluation, the output results are resized to match the original ground truth shapes, following which metrics are computed.

In the following parts, we evaluate all methods based on several commonly used metrics.
\textit{Global Intersection over Union Rate} ($IoU$) is used to evaluate the pixel-level capacity of differring.
For total $N$ images, it is defined as:
\begin{equation}
    IoU = \frac{\sum_{i=1}^{N} TP_{i}}{\sum_{i=1}^{N} (GP_{i} + FP_{i})} ,
\end{equation}
where $TP_i$, $GP_i$, and $FP_i$ denote the count of true positive, ground truth positive, and false positive pixels of the $i$-th image, respectively.
\textit{Normalized Intersection over Union} ($nIoU$) is used to evaluate the generalization ability across different scenarios, defined as:
\begin{equation}
    nIoU = \frac{1}{N} \sum_{i=1}^{N} \frac{TP_{i}}{GP_{i} + FP_{i}} .
\end{equation}

\textit{Global False Alarm Rate} ($F_a$) signifies the ratio of falsely predicted pixels to all the pixels in the image, and is defined as:
\begin{equation}
    F_a = \frac{\sum_{i=1}^{N} FP_{i}}{\sum_{i=1}^{N} (W_{i} \times H_{i})} ,
\end{equation}
where $W_i$ and $H_i$ denote the width and height of the $i$-th image, respectively.
\textit{Precision of Target Detection} ($P_d$) is a target-level metric utilized for evaluating the capability to distinguish objects, and is defined as:
\begin{equation}
    P_d = \frac{\sum_{i=1}^{N} \widehat{D}_i}{\sum_{i=1}^{N} D_i} ,
\end{equation}
where $\widehat{D}_i$ and $D_i$ denote the estimated and actual count of targets in the $i$-th image, respectively.
We employ the Moore neighborhood strategy to establish pixel-wise associations.

\subsection{Ablation Study} \label{subsec:hintonly}

\begin{figure}[t]
    \centering
    \includegraphics[width=1\linewidth]{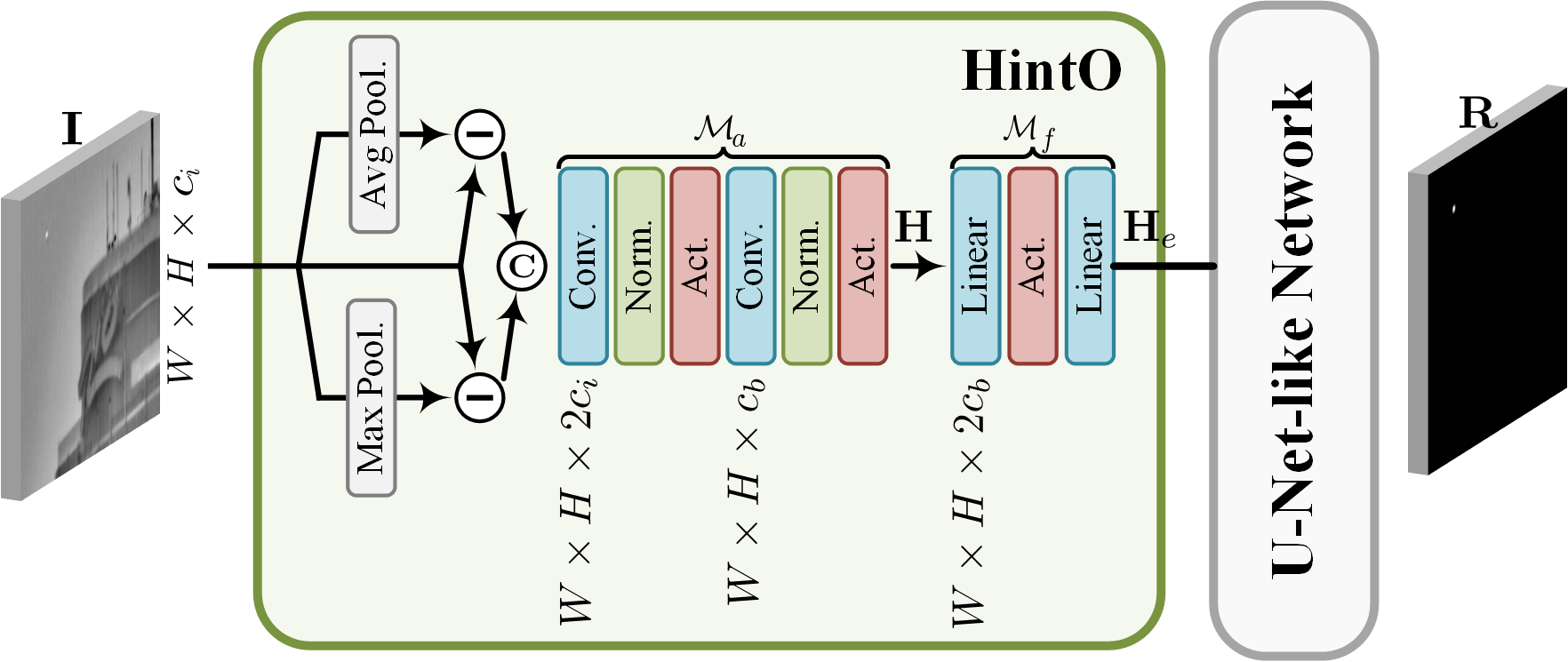}
    \caption{
    An overview of the HintO for infrared small target detection.
    In comparison to HintU, HintO completely excludes the original image when generating Hint representation, facilitating a direct assessment of the effectiveness of the Hint concept itself.}
    \label{fig:ab-network-hinto}
\end{figure}

\begin{table}[th]
    \caption{Comparisons of UNet/MiM+ with +HintO on NUDT-SIRST, SIRSTv2, and IRSTD1K.
    The rows using ``HintO'' are highlighted in \colorbox[RGB]{211,211,211}{light gray}.
    \textcolor[RGB]{150,25,25}{Red font} indicates performance degradation, while \textcolor[RGB]{25,150,25}{green font} indicates improvement.
    } \label{tab-hinto}
    \centering
    \begin{tabular}{cl||cccc}
    \toprule
    \multirow{2}{*}{Dataset}& \multirow{2}{*}{Method}& IoU& nIoU& $F_a$  & $P_d$ \\
 & & (\%)& (\%)& ($10^{-6}$)&(\%)\\
    \midrule
    \multirow{6}{*}{NUDT-SIRST} 
    & MiM+   & 63.57 & 63.00 & 11.14 & 89.65 \\
    & \cellcolor[HTML]{D3D3D3}+HintO & \cellcolor[HTML]{D3D3D3}65.09 & \cellcolor[HTML]{D3D3D3}63.73 & \cellcolor[HTML]{D3D3D3}4.55  & \cellcolor[HTML]{D3D3D3}89.87  \\
    &        & \cellcolor[HTML]{D3D3D3}{\color[HTML]{199619} +1.51} & \cellcolor[HTML]{D3D3D3}{\color[HTML]{199619} +0.73} & \cellcolor[HTML]{D3D3D3}{\color[HTML]{199619} -6.59} & \cellcolor[HTML]{D3D3D3}{\color[HTML]{199619} +0.23} \\
    & UNet   & 88.65 & 89.76 & 15.17 & 98.38 \\
    & \cellcolor[HTML]{D3D3D3}+HintO & \cellcolor[HTML]{D3D3D3}90.34 & \cellcolor[HTML]{D3D3D3}90.92 & \cellcolor[HTML]{D3D3D3}8.93  & \cellcolor[HTML]{D3D3D3}98.83 \\
    &        & \cellcolor[HTML]{D3D3D3}{\color[HTML]{199619} +1.69} &\cellcolor[HTML]{D3D3D3} {\color[HTML]{199619} +1.16} & \cellcolor[HTML]{D3D3D3}{\color[HTML]{199619} -6.24} & \cellcolor[HTML]{D3D3D3}{\color[HTML]{199619} +0.45}\\
    \midrule
    \multirow{6}{*}{SIRSTv2}
    & MiM+   & 53.77 & 55.54 & 28.76 & 88.45 \\
    & \cellcolor[HTML]{D3D3D3}+HintO & \cellcolor[HTML]{D3D3D3}53.87 & \cellcolor[HTML]{D3D3D3}55.58 & \cellcolor[HTML]{D3D3D3}21.57 & \cellcolor[HTML]{D3D3D3}89.59 \\
    &        & \cellcolor[HTML]{D3D3D3}{\color[HTML]{199619} +0.10} & \cellcolor[HTML]{D3D3D3}{\color[HTML]{199619} +0.05} & \cellcolor[HTML]{D3D3D3}{\color[HTML]{199619} -7.19} & \cellcolor[HTML]{D3D3D3}{\color[HTML]{961919} -1.14} \\
    & UNet   & 62.82 & 67.86 & 53.26 & 88.48 \\
    & \cellcolor[HTML]{D3D3D3}+HintO & \cellcolor[HTML]{D3D3D3}63.06 & \cellcolor[HTML]{D3D3D3}67.91 & \cellcolor[HTML]{D3D3D3}16.28 & \cellcolor[HTML]{D3D3D3}89.67 \\
    &        & \cellcolor[HTML]{D3D3D3}{\color[HTML]{199619} +0.24} & \cellcolor[HTML]{D3D3D3}{\color[HTML]{199619} +0.06} & \cellcolor[HTML]{D3D3D3}{\color[HTML]{199619} -36.98} & \cellcolor[HTML]{D3D3D3}{\color[HTML]{199619} +1.19} \\
    \midrule
    \multirow{6}{*}{IRSTD1K}
    & MiM+   & 52.31 & 47.86 & 16.45 & 78.60 \\
    & \cellcolor[HTML]{D3D3D3}+HintO & \cellcolor[HTML]{D3D3D3}54.08 & \cellcolor[HTML]{D3D3D3}47.95 & \cellcolor[HTML]{D3D3D3}24.83 & \cellcolor[HTML]{D3D3D3}76.08 \\
    &        & \cellcolor[HTML]{D3D3D3}{\color[HTML]{199619} +1.77} & \cellcolor[HTML]{D3D3D3}{\color[HTML]{199619} +0.09} & \cellcolor[HTML]{D3D3D3}{\color[HTML]{961919} +8.39} & \cellcolor[HTML]{D3D3D3}{\color[HTML]{961919} -2.52} \\
    & UNet   & 58.83 & 55.68 & 29.88 & 90.20 \\
    & \cellcolor[HTML]{D3D3D3}+HintO & \cellcolor[HTML]{D3D3D3}61.01 & \cellcolor[HTML]{D3D3D3}55.93 & \cellcolor[HTML]{D3D3D3}13.58 & \cellcolor[HTML]{D3D3D3}88.56 \\
    &        & \cellcolor[HTML]{D3D3D3}{\color[HTML]{199619} +2.18} & \cellcolor[HTML]{D3D3D3}{\color[HTML]{199619} +0.25} & \cellcolor[HTML]{D3D3D3}{\color[HTML]{199619} -16.30} & \cellcolor[HTML]{D3D3D3}{\color[HTML]{961919} -1.64} \\
    \bottomrule
    \end{tabular} 
\end{table}

It is unjust to assess the effectiveness of Hint only as a plug-in.
Despite its lightweight nature, it introduces some additional parameters, making it uncertain whether it will effectively work as our expected mechanism.
To verify it, we design the HintO (Hint Only) experiment, whose structure is depicted in Figure~\ref{fig:ab-network-hinto}.
In brief, in HintO, the original image is now totally excluded from the Hint.
We use both UNet and MiM+ as the baseline for their light weight, which might be more obvious when observing the effects of HintO.
The final results of MiM+ with HintO are presented in Table~\ref{tab-hinto}.
There is no doubt that the HintO strategy improves the pixel-level performance, and both UNet and MiM+ show performance improvements on all three datasets.
The poor performance of $P_d$ may be caused by insufficient parameters in MiM+ and UNet, as it is significantly more pronounced in MiM+.

In Figure~\ref{fig:ab-vis-hinto}, we present the feature representations of each step in HintO for a detailed analysis of its impact.
Subfigures (b-1) and (b-2) validate the hypothesis outlined in Equation~\ref{eq-hprime}.
Based on it, the $\mathcal{M}_a$ infers a coarse area potentially containing the targets, depicted as subfiture (c).
Following the final MLP $\mathbf{M}_f$, the hints coalesce more tightly.
While visually, $\mathbf{H}_e$ appears less significant than $\mathbf{H}$, in practice, they mark a more concentrated region of high-response cues.
Subsequent processing by the UNet involves straightforward denoising of a comparatively clear feature representation, a task considerably simpler than discerning targets amidst heavy image interference.

\begin{figure}[t]
    \centering
    \includegraphics[width=1.0\linewidth]{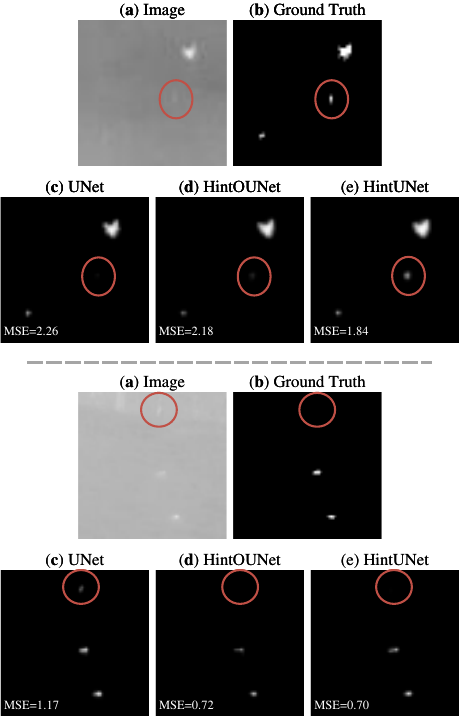}
    \caption{
    The distribution of target pixel ratios within each dataset relative to the entire image.
    The x-axis denotes the ratio of the number of target pixels to the total pixels within the respective frame.
    The y-axis shows the percentage of targets that correspond to each ratio range within the overall target count.}
    \label{fig:ab-hinto}
\end{figure}

Figure~\ref{fig:ab-hinto} shows two visualization examples exhibiting the improvements introduced by HintO and HintU to the vanilla UNet architecture.
In the upper one, a missed detection is evident.
UNet exhibits a minimal response to the target highlighted within the red circle, whereas HintO demonstrates a faint response, and HintU successfully detects the target as expected.
The lower example illustrates a false detection.
UNet incorrectly identifies a non-existent target within the red circle, a alleviation significantly mitigated by both HintO and HintU.
Due to the small size of the cut patch, stretching during the visualization process may have impacted the visual effects.
We quantify the Mean Squared Error (MSE) for the regional original data and annotate it in the corresponding images.

The metrics and examples verify that the Hint mechanism is capable of learning distinct information from the main method and simultaneously offering robust supplementary information for HintU.

\subsection{Comparative Experiments} \label{subsec:comp_exp}

In order to verify the universality of HintU, we implement it based on UNet~\cite{unet}, UNet++~\cite{unet++}, UIUNet~\cite{uiunet}, MiM-ISTD~\cite{mimistd} and HCFNet~\cite{hcfnet}.
Four additional representative deep learning methods are selected for comparison, i.e., MDvsFA~\cite{mdvsfa}, ACM~\cite{acm}, ALCNet~\cite{alcnet} and DNANet~\cite{dnanet/nudtsirst}.
Table~\ref{tab-comp} presents a comprehensive comparison of various advanced methods on three datasets: NUDT-SIRST, SIRSTv2, and IRSTD1K. Each method's performance is evaluated based on several metrics: IoU (Intersection over Union), nIoU (normalized IoU), $F_a$ (False alarm rate), and $P_d$ (Probability of detection).
The table is divided into sections, with each section dedicated to a specific method.
Within each section, the performance metrics for the method are listed across the three datasets.
Highlighted rows in light gray indicate the incorporation of the ``Hint" strategy, which is compared against the baseline method.
Performance improvements are highlighted in green font, while performance degradation is highlighted in red font.
The best performance for each metric is emphasized with bold underline, while the second-best performance is marked with wavy underline.

The evaluation of advanced deep learning methods on the NUDT-SIRST, SIRSTv2, and IRSTD1K datasets reveals some interesting insights.
The NUDT-SIRST dataset shows that UNet++1024c is the best performer.
It has the highest IoU (94.39\%) and nIoU (94.35\%) with the second highest $F_a$ (3.91), which means it is very good at finding targets.
For the SIRSTv2 dataset, HCFNet with Hint achieves the highest IoU (70.17\%) and nIoU (71.67\%), while UIUNet with Hint stands out with substantial enhancements in detection probability ($P_d$), reaching 92.71\%.
Almost all methods show a significant decrease in $F_a$.
Lastly, on the IRSTD1K dataset, UIUNet with Hint does better in IoU (63.85\%) and nIoU (59.49\%), while vanilla UIUNet does better in detection probability ($P_d$) at 96.64\%.
The significant improvements demonstrate that the Hint mechanism is fit to all of the 3 selected datasets.
Among all the methods, UNet has the simplest structure and greatest potential (showing improvements in all metrics on all datasets).
This suggests that the Hint mechanism may essentially be an equivalence of some complex structures from other methods.

\begin{figure*}
    \centering
    \includegraphics[width=1\linewidth]{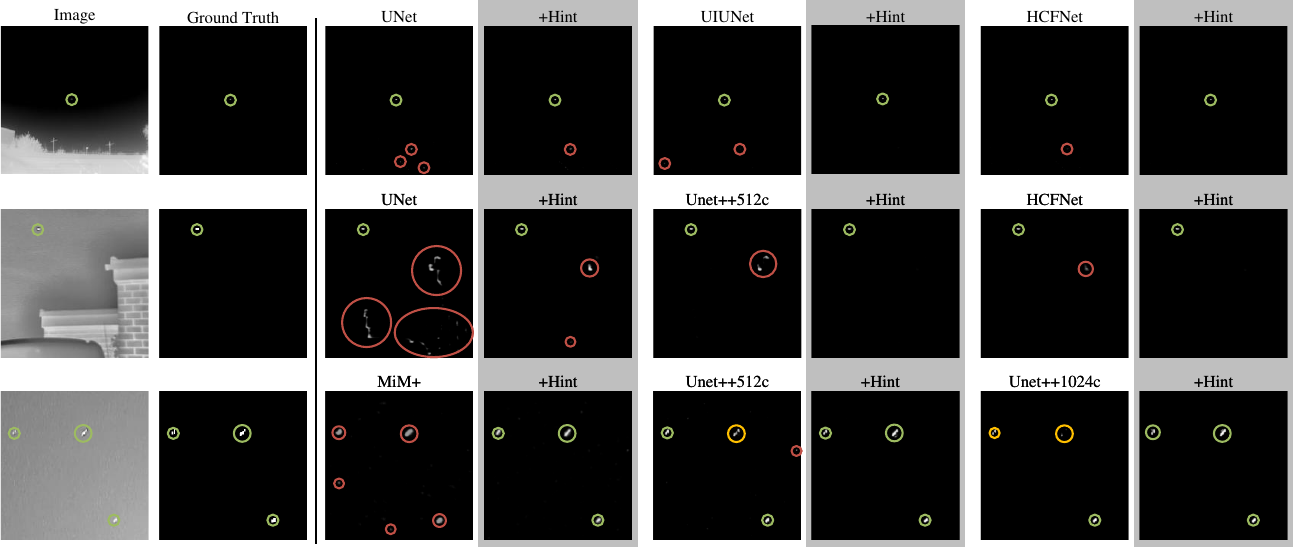}
    \caption{
    Visual comparison of the output results of different methods before and after applying HintU.
    Correct detections are denoted by \textcolor[RGB]{25,150,25}{green circles}, while false detections are highlighted with \textcolor[RGB]{150,25,25}{red circles}.
    Missed detections are represented by \textcolor[RGB]{255,192,0}{yellow circles}.
    Additionally, subfigures shaded in \colorbox[RGB]{211,211,211}{light gray} signify the application of the HintU.
    }
    \label{fig-reults-comp}
\end{figure*}

\begin{table*}[t]
    \caption{Comparisons with other advanced methods on NUDT-SIRST, SIRSTv2, and IRSTD1K.
    The ``Hint'' strategy is denoted by the rows highlighted in \colorbox[RGB]{211,211,211}{light gray}.
    \textcolor[RGB]{150,25,25}{Red font} indicates performance degradation, while \textcolor[RGB]{25,150,25}{green font} indicates improvement.
    The best performance of each metrics is highlighted with \textbf{\underline{bold underline}}, and the second-best is marked with \uwave{wavy underline}.
    } \label{tab-comp}
    \centering
    \setlength{\tabcolsep}{2.8mm}{}
    \begin{tabular}{l||cccc|cccc|cccc}
    \toprule
  \multirow{3}{*}{Method} & \multicolumn{4}{c}{\textbf{NUDT-SIRST}} & \multicolumn{4}{c}{\textbf{SIRSTv2}} & \multicolumn{4}{c}{\textbf{IRSTD1K}}\\
     & IoU & nIoU & $F_a$  & $P_d$  & IoU& nIoU& $F_a$  &$P_d$  & IoU& nIoU& $F_a$  &$P_d$  
\\
  & (\%)& (\%)& ($10^{-6}$)&(\%) & (\%)& (\%)& ($10^{-6}$)&(\%)  &  
(\%)& (\%)& ($10^{-6}$)&(\%)  \\
    \midrule
MDvsFA \cite{mdvsfa}    & 73.51 & 72.24 & 35.27 & 89.13 & 48.46 & 46.38 & 40.78 & 70.43 & 49.50 & 47.41 & 80.33 & 82.11 \\
ACM \cite{acm}       & 79.08 & 81.09 & 10.34 & 96.51 & 59.47 & 54.39 & 47.11 & 70.86 & 58.63 & 54.31 & 31.78 & 82.58 \\
ALCNet \cite{alcnet}    & 81.59 &  82.21 & 13.52  & 96.55 & 61.09 & 63.13 & 31.52 & 84.08 & 60.21 & 55.77 & 21.56 & 87.18 \\
DNANet \cite{dnanet/nudtsirst}     & 85.45 & 85.56 & 7.79  & 96.43 & 64.93 & 66.98 & 23.58 & 91.49 & \uwave{63.15} & 56.77 & 32.41 & 91.67 \\
\midrule
MiM+$^\dagger$       & 63.57 & 63.00 & 11.14 & 89.65 & 53.77 & 55.54 & 28.76 & 88.45 & 52.31 & 47.86 & 16.45 & 78.60 \\
\rowcolor[HTML]{D3D3D3} 
+Hint (\textbf{Ours})& 66.55 & 66.63 & 27.66 & 93.17 & 61.36 & 65.02 & 15.27 & 89.67 & 56.46 & 52.69 & 18.86 & 83.71 \\
\rowcolor[HTML]{D3D3D3} 
& \textcolor[RGB]{25,150,25}{+2.98} & \textcolor[RGB]{25,150,25}{+3.63} &	\textcolor[RGB]{150,25,25}{+16.52} &	\textcolor[RGB]{25,150,25}{+3.52} &	\textcolor[RGB]{25,150,25}{+7.59} &	\textcolor[RGB]{25,150,25}{+9.48} &	\textcolor[RGB]{25,150,25}{-13.49} &	\textcolor[RGB]{25,150,25}{+1.22}	& \textcolor[RGB]{25,150,25}{+4.15} &	\textcolor[RGB]{25,150,25}{+4.83} &	\textcolor[RGB]{150,25,25}{+2.41} &	\textcolor[RGB]{25,150,25}{+5.11}\\
\midrule
Unet~\cite{unet}     & 88.65 & 89.76 & 15.17 & 98.38 & 62.82 & 67.86 & 53.26 & 88.48 & 58.83 & 55.68 & 29.88 & 90.20 \\
\rowcolor[HTML]{D3D3D3} 
\rowcolor[HTML]{D3D3D3} 
+Hint (\textbf{Ours})& 92.49 & 92.61 & 7.30  & 98.83 & 65.40 & 69.93 & 30.23 & 89.06 & 62.25 & 57.32 & 26.37 & 91.12 \\
\rowcolor[HTML]{D3D3D3} 
& \textcolor[RGB]{25,150,25}{+3.84} & \textcolor[RGB]{25,150,25}{+2.86} &	\textcolor[RGB]{25,150,25}{-7.87} &	\textcolor[RGB]{25,150,25}{+0.45} &	\textcolor[RGB]{25,150,25}{+2.58} &	\textcolor[RGB]{25,150,25}{+2.07} &	\textcolor[RGB]{25,150,25}{-23.03} &	\textcolor[RGB]{25,150,25}{+0.58}	& \textcolor[RGB]{25,150,25}{+3.42} &	\textcolor[RGB]{25,150,25}{+1.64} &	\textcolor[RGB]{25,150,25}{-3.51} &	\textcolor[RGB]{25,150,25}{+0.92}\\
\midrule
Unet++512c~\cite{unet++}     & 91.53 & 92.18 & 7.91  & 98.48 & 62.61 & 67.74 & 40.19 & 86.72 & 59.75 & 55.94 & 33.58 & 91.73 \\
\rowcolor[HTML]{D3D3D3} 
+Hint (\textbf{Ours})& 91.93 & 92.45 & 4.81  & 98.61 & 64.38 & 68.46 & 20.51 & 89.67 & 61.82 & 57.24 & 22.13 & 90.85 \\
\rowcolor[HTML]{D3D3D3} 
& \textcolor[RGB]{25,150,25}{+0.41} & \textcolor[RGB]{25,150,25}{+0.27} &	\textcolor[RGB]{25,150,25}{-3.10} &	\textcolor[RGB]{25,150,25}{+0.13} &	\textcolor[RGB]{25,150,25}{+1.77} &	\textcolor[RGB]{25,150,25}{+0.73} &	\textcolor[RGB]{25,150,25}{-19.67} &	\textcolor[RGB]{25,150,25}{+2.94}	& \textcolor[RGB]{25,150,25}{+2.07} &	\textcolor[RGB]{25,150,25}{+1.30} &	\textcolor[RGB]{25,150,25}{-11.45} &	\textcolor[RGB]{150,25,25}{-0.89}\\
Unet++1024c     & \uwave{93.53} & \uwave{93.82} & 5.39 & 98.29 & 64.12 & 70.51 & 24.04 & 89.06 & 60.43 & 56.33 & \textbf{\underline{12.91}} & 89.10 \\
\rowcolor[HTML]{D3D3D3} 
+Hint (\textbf{Ours})& \textbf{\underline{94.39}} & \textbf{\underline{94.35}} & \uwave{3.91}  & 98.83 & 68.36 & \uwave{71.08} & \textbf{\underline{11.28}} & 90.58 & 60.53 & 57.20 & 19.50 & 90.58 \\
\rowcolor[HTML]{D3D3D3} 
& \textcolor[RGB]{25,150,25}{+0.87} & \textcolor[RGB]{25,150,25}{+0.53} &	\textcolor[RGB]{25,150,25}{-1.47} &	\textcolor[RGB]{25,150,25}{+0.53} &	\textcolor[RGB]{25,150,25}{+4.24} &	\textcolor[RGB]{25,150,25}{+0.57} &	\textcolor[RGB]{25,150,25}{-12.75} &	\textcolor[RGB]{25,150,25}{+1.52}	& \textcolor[RGB]{25,150,25}{+0.10} &	\textcolor[RGB]{25,150,25}{+0.87} &	\textcolor[RGB]{150,25,25}{+6.59} &	\textcolor[RGB]{25,150,25}{+1.48}\\
\midrule
UIUNet\cite{uiunet}     & 89.09 & 90.14 & 5.55  & 98.40 & 67.16 & 67.03 & \uwave{14.19} & 88.22 & 61.57 & 56.99 & 23.20 & \textbf{\underline{96.64}} \\
\rowcolor[HTML]{D3D3D3} 
+Hint (\textbf{Ours})& 91.85 & 92.38 & 4.90  & \textbf{\underline{99.36}} & 69.39 & 70.62 & 18.18 & \textbf{\underline{92.71}} & \textbf{\underline{63.85}} & \textbf{\underline{59.48}} & 16.17 & \uwave{93.14} \\
\rowcolor[HTML]{D3D3D3} 
& \textcolor[RGB]{25,150,25}{+2.76} & \textcolor[RGB]{25,150,25}{+2.24} &	\textcolor[RGB]{25,150,25}{-0.64} &	\textcolor[RGB]{25,150,25}{+0.96} &	\textcolor[RGB]{25,150,25}{+2.23} &	\textcolor[RGB]{25,150,25}{+3.59} &	\textcolor[RGB]{150,25,25}{+3.99} &	\textcolor[RGB]{25,150,25}{+4.48}	& \textcolor[RGB]{25,150,25}{+2.28} &	\textcolor[RGB]{25,150,25}{+2.49} &	\textcolor[RGB]{25,150,25}{-7.03} &	\textcolor[RGB]{150,25,25}{-3.50}\\
\midrule
HCFNet~\cite{hcfnet}     & 90.65 & 90.71 & 4.83  & 98.51 & \uwave{69.41} & 70.31 & 19.38 & 89.97 & 60.63 & 56.06 & \uwave{15.11} & 88.23 \\
\rowcolor[HTML]{D3D3D3} 
+Hint (\textbf{Ours})& 93.29 & 93.67 & \textbf{\underline{0.67}}  & \uwave{98.93} & \textbf{\underline{70.17}} & \textbf{\underline{71.67}} & 17.72 & \uwave{91.79} & 62.24 & \uwave{58.87} & 17.12 & 90.17\\
\rowcolor[HTML]{D3D3D3} 
& \textcolor[RGB]{25,150,25}{+2.64} & \textcolor[RGB]{25,150,25}{+2.96} &	\textcolor[RGB]{25,150,25}{-4.16} &	\textcolor[RGB]{25,150,25}{+0.42} &	\textcolor[RGB]{25,150,25}{+0.76} &	\textcolor[RGB]{25,150,25}{+1.36} &	\textcolor[RGB]{25,150,25}{-1.66} &	\textcolor[RGB]{25,150,25}{+1.82}	& \textcolor[RGB]{25,150,25}{+1.61} &	\textcolor[RGB]{25,150,25}{+2.81} &	\textcolor[RGB]{150,25,25}{+2.01} &	\textcolor[RGB]{25,150,25}{+1.94}\\
    \bottomrule
 \multicolumn{12}{l}{$^\dagger$ A lightweight variant of MiM-ISTD \cite{mimistd}.}
    \end{tabular} 
\end{table*}

\subsection{Efficiency Analysis} \label{subsec-effexp}

Table~\ref{tab-efficiency-comp} provides a comparison of computational costs for all mentioned methods in Table~\ref{tab-comp}.
The evaluation metrics include the number of floating point operations (GFLOPs, $\times10^9$ FLOPs), the number of trainable parameters (MParas, $\times10^6$ Parameters), and frame-average inference time (ms, based on 1000 frames, excluding the time to load from IO).
Due to the gap in GPU performance, inference time is a relative reference, and we provide the normalized frame-average inference time (ms\%, using HCFNet+Hint as the 100\% benchmark) and the additional cost of the Hint module (+ms\%, measuring the percentage of additional inference time caused by the Hint module in each method).
As the parameter quantity of Hint is entirely under control, the increment in parameter quantity is minimal, with an average increment of approximately 0.27 million.
Due to Hint's reliance on poolings, although there is no significant increase in parameter count, it results in a higher number of FLOPs.
Specifically, among all methods implementing Hint, an average increase of 23.05 GFLOPs is observed, consequently leading to longer computation times, with an average increment of 1.88ms across all methods.
Relative to the original method without Hint, the average inference time experiences a 29.47\% increase.
However, this increase is relatively modest for methods like MiM+, HCFNet, and UIUNet, which inherently incur high inference time costs.
This can be attributed to Hint's core being completely independent of the main method, thereby maintaining consistent complexity regardless of changes in it.

\begin{table}[th]
    \caption{
     A comparison of computational costs for all mentioned methods in Table~\ref{tab-comp}.
     The ``Hint'' strategy is denoted by the rows highlighted in \colorbox[RGB]{211,211,211}{light gray}.
    } \label{tab-efficiency-comp}
    \centering
    \begin{tabular}{l||ccccc}
    \toprule
Method          & GFLOPs  & MParas & ms    & ms\%   & +ms\%  \\
\midrule
ACM             & 0.28   & 0.39  & 5.50  & 12.26  & -      \\
ALCNet          & 0.45   & 0.52  & 7.50  & 16.72  & -      \\
DNANet          & 14.06  & 4.70  & 20.72 & 46.20  & -      \\
\midrule
MiM+            & 0.59   & 1.93  & 22.03 & 49.12  & -      \\
\rowcolor[HTML]{D3D3D3} 
+Hint           & 22.87  & 2.24  & 23.53 & 52.47  & 6.83   \\
\midrule
UNet++512c      & 34.62  & 9.16  & 4.42  & 9.85   & -      \\
\rowcolor[HTML]{D3D3D3} 
+Hint           & 57.45  & 9.48  & 6.35  & 14.15  & 43.61  \\
\midrule
UNet            & 48.16  & 31.03 & 2.79  & 6.21   & -      \\
\rowcolor[HTML]{D3D3D3} 
+Hint           & 71.58  & 31.36 & 4.71  & 10.50  & 69.08  \\
\midrule
UNet++1024c     & 138.09 & 36.63 & 4.42  & 9.85   & -      \\
\rowcolor[HTML]{D3D3D3} 
+Hint           & 161.51 & 36.95 & 6.34  & 14.14  & 43.52  \\
\midrule
UIUNet          & 54.39  & 50.54 & 21.52 & 47.99  & -      \\
\rowcolor[HTML]{D3D3D3} 
+Hint           & 77.81  & 50.87 & 23.41 & 52.18  & 8.74   \\
\midrule
HCFNet          & 21.80  & 15.29 & 42.71 & 95.22  & -      \\
\rowcolor[HTML]{D3D3D3} 
+Hint           & 44.70  & 15.61 & 44.85 & 100.00 & 5.02   \\
\bottomrule
    \end{tabular} 
\end{table}

\subsection{Further Explorations} \label{subsec-principlesexp}

\begin{figure*}[t]
    \centering
    \includegraphics[width=1\linewidth]{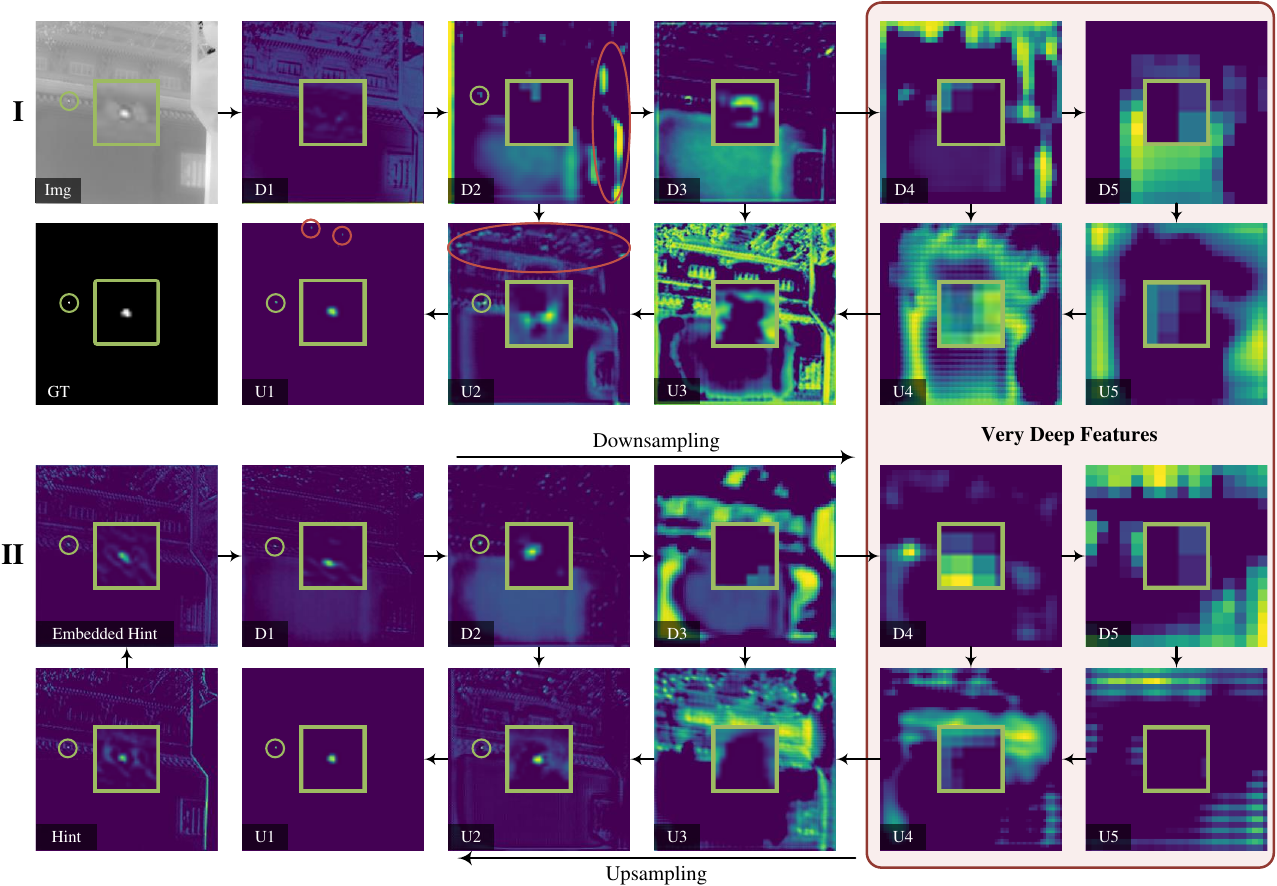}
    \caption{
    Using UNet as an example, we illustrate a comparative analysis of feature flow before and after the Hint mechanism is applied.
    The upper two rows depict UNet, while the lower two rows illustrate UNet+HintU.
    Feature flows within the image are delineated with arrows based on UNet's actual operations.
    We use \colorbox[RGB]{100,100,100}{\textcolor{white}{white text on a black background}} at the bottom left of each sub-figure to denote the status, corresponding the states in Figure~\ref{fig-unet-comp}-(\textbf{a}).
    As \colorbox[RGB]{100,100,100}{\textcolor{white}{D4}}, \colorbox[RGB]{100,100,100}{\textcolor{white}{D5}}, \colorbox[RGB]{100,100,100}{\textcolor{white}{U4}}, and \colorbox[RGB]{100,100,100}{\textcolor{white}{U5}} are located within the very deep layers of UNet, their features have achieved a high level of abstraction, making them almost disunderstandable.
    For enhanced visibility, highlighted areas are enlarged.
    }
    \label{fig-viscomp}
\end{figure*}

\begin{figure*}
    \centering
    \includegraphics[width=1\linewidth]{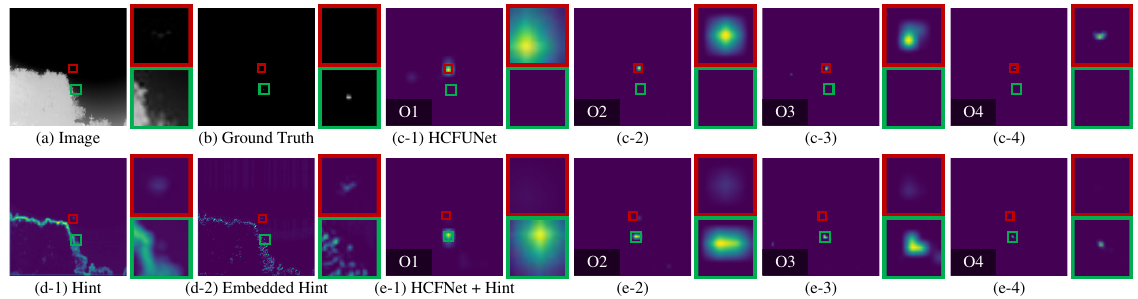}
    \caption{
    Visual comparison of different layers on HCFNet and HCFNet+Hint in an extreme example.
    The target is denoted by \textcolor[RGB]{25,150,25}{green rectangles}, while a distractor is highlighted with \textcolor[RGB]{150,25,25}{red rectangles}.
    We use \colorbox[RGB]{100,100,100}{\textcolor{white}{white text on a black background}} at the bottom left of some sub-figures to denote the status, corresponding the states in Figure~\ref{fig-unet-comp}-(\textbf{b}).
    The state \colorbox[RGB]{100,100,100}{\textcolor{white}{O4}} is also the final result of the methods.
    For enhanced visibility, highlighted areas are enlarged.
}
    \label{fig-hcf-vis}
\end{figure*}

\begin{figure}[t]
    \centering
    \includegraphics[width=1\linewidth]{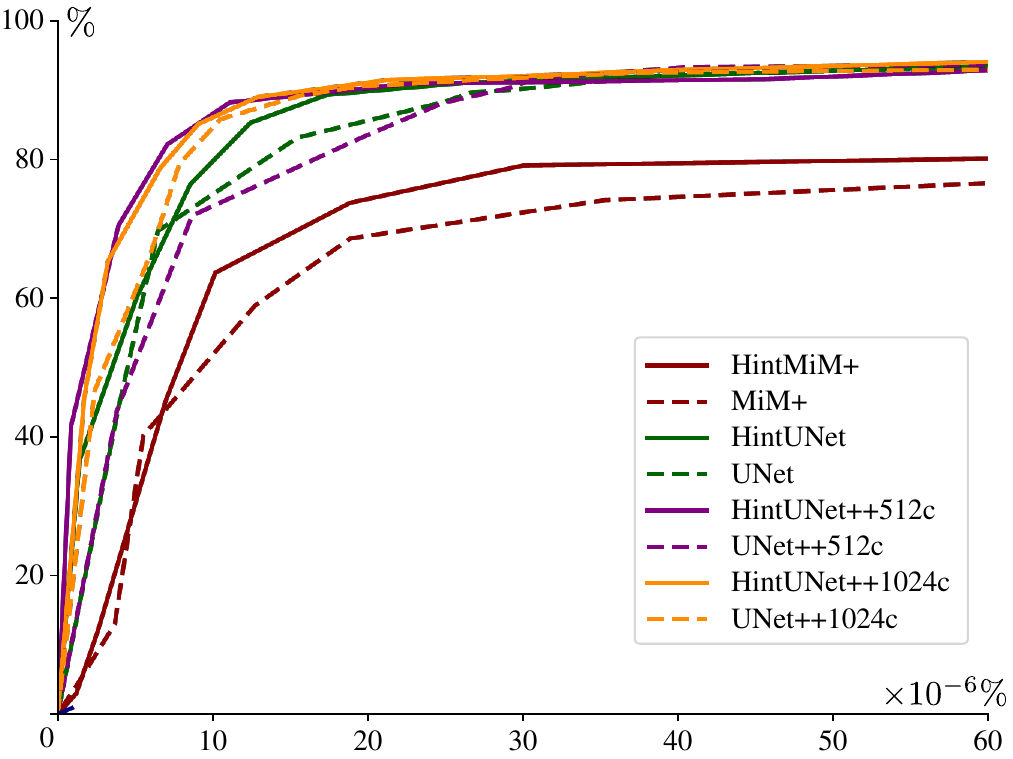}
    \caption{
    The $P_d$-$F_a$ curve of MiM+, UNet, UNet++, and their Hint version.
    The x-axis denotes the ratio of the number of target pixels to the total pixels within the respective frame.
    The y-axis shows the percentage of targets that correspond to each ratio range within the overall target count.}
    \label{fig-roccurve}
\end{figure}

The introduction of the Hint mechanism has a significant and stable improvement in IoU and nIoU, but it may lead to a certain degree of decline in $F_a$ and $P_d$.
Therefore, we draw a curve graph for the $F_a$-$P_d$ relationship of some methods, as shown in Figure~\ref{fig-roccurve}.
The dotted line in the figure is the original method, and the solid line with the same color represents the method after applying Hint.
The figure shows that within the acceptable $F_a$ range, the overall curve of the improved method is better than its original version.
This also explains why often only one of the two will experience the performance degradation. In essence, it is just a difference in selection points, rather than an overall performance degradation.

Besides the universality, there is another highlight in Table~\ref{tab-comp}.
It is important to highlight that while UNet boasts a simplistic architecture, it typically falls short when compared to methods designed for ISTD.
However, leveraging the HintU can lead to notable performance enhancements, rendering it competitive with other methods, including those also employing HintU.
Our curiosity is attracted by exploring the underlying workflow inherent to its remarkable effects.

Figure~\ref{fig-viscomp} presents a visual experiment of the feature dynamics during all the intermediate states of processing in UNet and its HintU variant.
In the vanilla UNet, the most shallow D1 layer hardly captures any valuable information of the target of interest.
While the D2 layer exhibits a faint and ambiguous response to the target, it is accompanied by numerous false-positive pixels that overshadow the target's significance.
An analysis of the entire process reveals that from the D3 layer to the U3 layer, the feature representations are very abstract.
Particularly noteworthy is the very deep-level nature of the D4, D5, U4, and U5 layers, which pose challenges for maintaining shallow semantics.
This abstraction poses a significant obstacle for the detection of ultra-tiny targets, as evident in the context of ISTD.
Unfortunately, this interference affects not only the downsampling process but also the upsampling layers via the skip connection, thereby perpetuating a lot of false alarm information.
The aforementioned deficiencies lead to false detections in the final output (U1).

In contrast, the initial Hint in the HintU promptly marks the target.
Take a closer look at Hint in HintU and the D2 state in UNet.
In fact, they have similar patterns (D2 is resized from the $1/4$ size of Hint), which means that UNet takes two steps to infer information similar to the initial Hint.
This characteristic is further amplified in the embedded Hint layer, where the target's importance is markedly pronounced.
This enhancement culminates in output that is discernibly clearer and devoid of artifacts.
The visualization offers a glimpse into the underlying effects of HintU on the UNet.

This experiment verifies that HintU indeed serves as a priori reminder in UNet's reasoning flow, and that the reminder is accurately captured by UNet, effectively improving the reasoning effect of each state.
It is in line with our original design intention: to provide a "Hint" that is more suitable for extremely tiny targets.
This mechanism apparently quickly provides the network with effective prior information and exhibits an attention-like activation state.

Since UNet is optimized within end-to-end strategy, the intermediate features are completely uncontrollable, resulting in most layers in the visualization being unreadable.
Therefore, we also conduct an additional visualization experiment on HCFNet.
Benefiting from the deep supervision strategy used by HCFNet, each step of its upsampling can be represented as an output of different resolutions.
This not only forces the UNet structure to pay close attention to shallow semantics, but also provides excellent visualization and interpretability.
Figure~\ref{fig-hcf-vis} is a visual experiment of a very extreme scenario.
The target (marked by the green rectangle) is very close to the extremely high response area in the lower left corner and has a lower response value.
Additionally, a distractor (marked by a red rectangular box) appears in an area where it is easier to detect.
Even the human visual system is difficult to distinguish without knowing the ground truth.

The results of HCFNet are displayed in the upper row. Although the entire inference process appears reasonable and accurate, it slowly approaches the inference target hierarchically without too many false alarms.
But it completely ignores the underlying real goals and only considers the most obvious ones.
From the false alarms in the canopy area in (c-1), we can infer that HCFNet may tend to respond to high-response areas in the original image.

The lower column shows the results of HCFNet under the HintU framework. From (d-1), it can be found that the spatial scale of the target is enlarged, and the responsiveness of the distractor is reduced. The two have similar responsiveness. This Given a priori equality.
In (d-2), the target is accurately labeled and given an intensity even stronger than the distractor, probably because Hint is essentially learning a new, more discriminative latent space pattern.
In the layer-by-layer reasoning process (e), we can observe that in (e-2) and (e-3), the network also responded weakly to the distractor, but this false alarm was quickly suppressed suppression, and finally output the correct target detection result.

\subsection{Generalization Analysis} \label{subsec-generalizationexp}

\begin{table}[th]
    \caption{UIUNet and UIUNet+Hint are trained comprehensively on the entire NUDT-SIRST dataset.
    Subsequently, evaluations were conducted on the other two datasets to confirm the difference in generalization capacity between the two models.} \label{tab-gene}
    \centering
    \begin{tabular}{cc||cccc}
    \toprule
    \multirow{2}{*}{Dataset}& \multirow{2}{*}{Method}& IoU& nIoU& $F_a$  & $P_d$ \\
 & & (\%)& (\%)& ($10^{-6}$)&(\%)\\
    \midrule
    \multirow{6}{*}{IRSTD1K}
    & UIU 100E  & 41.98 & 49.73 & 145.62& 79.75 \\
    & UIU 200E  & 44.02 & 51.07 & 137.67& 84.42 \\
    & UIU 300E  & 42.51 & 48.21 & 141.70& 82.64  \\
    & Hint 100E & 40.61 & 43.54 & 173.25& 76.62 \\
    & Hint 200E & 43.77 & 48.33 & 99.88 & 83.31 \\
    & Hint 300E & 44.60 & 50.58 & 86.94 & 85.87 \\
    \midrule
    \multirow{6}{*}{SIRSTv2}
    & UIU 100E  & 58.72 & 53.86 & 76.52 & 80.96 \\
    & UIU 200E  & 61.45 & 55.71 & 36.73 & 91.25 \\
    & UIU 300E  & 61.57 & 56.99 & 33.20 & 90.64  \\
    & Hint 100E & 58.42 & 56.31 & 69.88 & 80.20 \\
    & Hint 200E & 60.37 & 57.64 & 31.12 & 87.59 \\
    & Hint 300E & 63.38 & 61.22 & 12.16 & 90.54 \\
    \bottomrule
    \end{tabular} 
\end{table}

Let's reconsider Hint from another perspective.
Hint introduces a priori guidance to the main methods, relying more on patterns than on data.
It offers a focus map indicating which pixels might be "interesting" based on the original image rather than on deep and abstract semantics.
This prompts us to question whether Hint could enhance the generalization capacity of primary methods, especially in scenarios involving different domains.
As depicted in Figure~\ref{fig:ds-size-dist}, the distributions of each dataset are dissimilar, pointing to distinct feature subspaces.
We conducted a new experiment training on the entire NUDT-SIRST dataset and testing on the other two datasets.
We chose UIUNet as the baseline due to its sufficient parameters to resist overfitting.
The results are presented in Table~\ref{tab-gene}.

In testing on the IRSTD1K dataset, UIUNet's performance at 200 epochs surpassed that at 300 epochs, suggesting that overfitting likely began between 200 and 300 epochs.
Given the number of ultra-tiny targets in NUDT-SIRST (less than 0.01\% of the full frame), significantly fewer than IRSTD1K, there's a risk of larger objects being forgotten during prolonged training on NUDT-SIRST.
For UIUNet+Hint, although the rate of performance improvement slows, there's no clear indication of overfitting.

The distribution of SIRSTv2 aligns more closely with NUDT-SIRST.
Consequently, UIUNet doesn't exhibit significant overfitting, but the performance gains from 200 to 300 epochs are marginal, likely due to reaching feature tolerance.
Nonetheless, UIUNet+Hint can still learn effective features and even achieves competitive performance compared to training with SIRSTv2.
This experiment validates Hint's contribution to generalization improvement and offers insights for future research endeavors.

\section{Conclusion}

This paper introduces HintU, a generic framework aimed at the ISTD task tailored for UNet-like networks, engineered to enrich the circular information commonly neglected by existing methods.
The core of HintU is a novel sub-prior information mechanism named ``Hint", which provides the pixels of interest to the network prior to inference, thereby augmenting its performance.
Notably, the integration of Hint brings only minimal computational overhead, approximately 2 milliseconds, but yields substantial performance improvements.
Importantly, the information embedded within Hint originates from the raw image data itself, representing a novel departure from conventional deep learning methodologies in image processing.
To validate the efficacy of Hint, we conduct HintO experiments, affirming that the information within Hint in isolation is capable of achieving competitive, if not the best, performance levels.
Subsequently, comprehensive experiments across different datasets, including NUDT-SIRST, SIRSTv2, and IRSTD1K, are undertaken to evaluate the effects of HintU across various architectures, including vanilla UNet, UNet++, UIUNet, MiM+, and HCFNet.
The experimental results demonstrate the substantial advantages conferred by HintU over these baseline methods.
Finally, a detailed exploration of some foundational principles and highlights underpinning Hint is conducted, in addition to its pivotal role in enhancing model generalization capabilities.

As technology progresses and societal landscapes evolve, the existence of small-sized objects within complex and unknown scenes becomes more prevalent.
Simply migrating generic image processing techniques to the domain of ISTD would inevitably miss a lot of inherant and specific characteristics, an outcome all researchers seek to avoid.
To enhance the robustness and adaptability of models to increasingly diverse and unknown scenarios, exploring further the unique characteristics of the ISTD task itself emerges as an important path for research.
This entails a deep understanding of the whole task, enabling the development of more specific methods that can effectively capture and leverage the inherent mechanisms of ISTD.

% \bibliography{ref.bib}

\end{document}